\documentclass[10pt,a4paper]{article}
\usepackage{natbib}
\usepackage{times}
\usepackage[font=small,labelfont=bf]{caption}
\newcommand{\institute}[1]{}
\date{To appear in {\em Encyclopedia of systems and control, 2nd edition}}

\usepackage{graphicx}    %
\usepackage{amsmath}
\usepackage{tikz}

\usepackage{color}

\begin{document}

\title{The Use of Gaussian Processes in System Identification}

\author{Simo S\"arkk\"a  %
}

\institute{Aalto University\\ %
simo.sarkka@aalto.fi         %
}

\maketitle

\section{Abstract}

Gaussian processes are used in machine learning to learn input-output mappings from observed data. Gaussian process regression is based on imposing a Gaussian process prior on the unknown regressor function and statistically conditioning it on the observed data. In system identification, Gaussian processes are used to form time series prediction models such as non-linear finite-impulse response (NFIR) models as well as non-linear autoregressive (NARX) models. Gaussian process state-space models (GPSS) can be used to learn the dynamic and measurement models for a state-space representation of the input-output data. Temporal and spatio-temporal Gaussian processes can be directly used to form regressor on the data in the time domain. The aim of this article is to briefly outline the main directions in system identification methods using Gaussian processes.

\section{Keywords}

Gaussian process regression, non-linear system identification, GP-NFIR model, GP-ARX model, GP-NOE model, Gaussian process state-space model, temporal Gaussian process, state-space Gaussian process

\section{Introduction}

Gaussian process regression \citep{Rasmussen+Williams:2006} refers to a statistical methodology where we use Gaussian processes as prior models for regression functions that we fit to observed data. This kind of methodology is particularly popular in machine learning although the origins of the basic ideas can be traced to geostatistics \citep{Cressie:1993}. In geostatistics, the corresponding methodology is called "kriging" which is named after South African mining engineer D.\ G.\ Krige. In system identification, Gaussian processes can be used to identify (or "learn" in machine learning terms) the input-output relationships from observed data. Even when there is no explicit input in the system, Gaussian processes can be used to identify a model for an observed time series of outputs which is a specific form of a system identification problem. Overviews of the use of Gaussian processes in system identification can be found, for example, in the monograph of \citet{Kocijan:2016}, and PhD theses of \citet{McHutchon:2014} and \citet{Frigola:2016}.

\section{Gaussian processes in system identification}

\subsection{Gaussian process regression}

Gaussian process regression is concerned with the following problem: Given a set of observed (training) input-output data $\mathcal{D} = \{ (\mathbf{z}_k, y_k) : k=1,\ldots,N \}$ from an unknown function $y = f(\mathbf{z})$, predict the values of the function at new (test) inputs $\{ \mathbf{z}_k^* : k=1,\ldots,M \}$. That is, the problem is a classical regression problem. However, the classical solution to the problem usually amounts to fixing a parametric function class $f(\mathbf{z};\boldsymbol{\theta})$, where $\boldsymbol{\theta}$ is a set of parameters and then fitting the parameters to the observed data. In Gaussian process regression, we take a different route; instead of fixing a parametric class of functions, we put a Gaussian process prior measure on the whole regression function and condition on the observed data using Bayes' rule \citep{Rasmussen+Williams:2006}. 

\subsubsection{Gaussian process regression problem}

Mathematically, the Gaussian process regression problem can be written as
\begin{subequations}
\begin{align}
  f(\mathbf{z}) &\sim \mathcal{GP}(m(\mathbf{z}), k(\mathbf{z},\mathbf{z}')), \label{eq:gp_prior} \\
  y_k &= f(\mathbf{z}_k) + \epsilon_k, \quad \epsilon_k \sim \mathcal{N}(0,\sigma^2_\mathrm{n}),
   \label{eq:gp_likelihood}
\end{align}
\end{subequations}
where Equation~\eqref{eq:gp_prior} tells that, a priori, the function is a Gaussian process with mean function $m(\mathbf{z}) = \mathrm{E}[f(\mathbf{z})]$ and covariance function (or kernel) $k(\mathbf{z},\mathbf{z}') = \mathrm{Cov}[f(\mathbf{z}),f(\mathbf{z'})] = \mathrm{E}[ (f(\mathbf{z}) - m(\mathbf{z})) \, (f(\mathbf{z}') - m(\mathbf{z}'))]$. Equation~\eqref{eq:gp_likelihood} tells that we observe the function values at points $\mathbf{z}_k$, $k=1,\ldots,N$ and that they are corrupted by (independent) Gaussian noises with variance $\sigma_\mathrm{n}^2$.

The mean and covariance functions define the regressor function class and they, or at least their parametric classes, need to be selected a priori. The mean function can typically be selected to be identically zero $m(\mathbf{z}) = 0$. The covariance function defines the smoothness properties of the functions, and a typical choice in machine learning is the squared exponential covariance function
\begin{equation}
  k(\mathbf{z},\mathbf{z}') = s^2 \, \exp\left( -\frac{\| \mathbf{z} - \mathbf{z}' \|^2}{2\ell^2} \right)
\label{eq:se_cov}
\end{equation}
which produces infinitely differentiable (i.e., analytic) regressor functions. The parameters $s$ and $\ell$ in the aforementioned covariance function define the magnitude and length scales of the regressor functions, respectively. Other common choices of covariance functions are, for example, the Mat\'ern class of covariance functions \citep{Matern:1960,Rasmussen+Williams:2006}.

\subsubsection{Gaussian process regression equations}

\begin{figure}[htb!]
\centering
\includegraphics{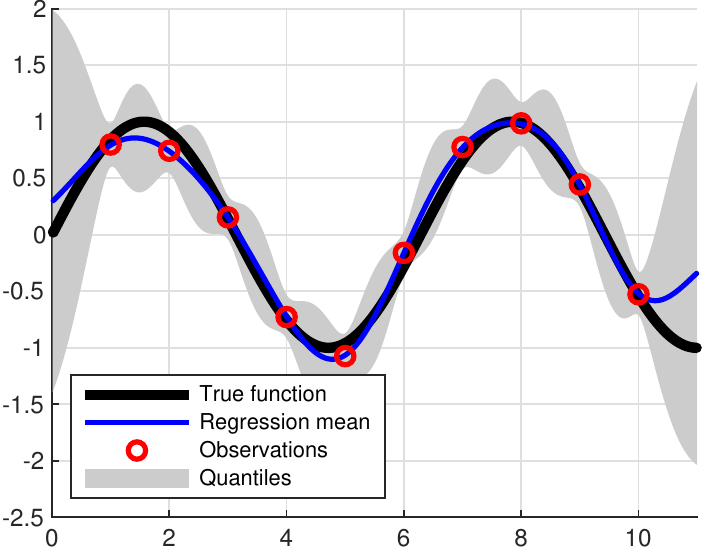}
\caption{Example of Gaussian process regression with squared exponential covariance function. The true function is a sinusoidal which is observed only at 10 points that are corrupted by Gaussian noise. The quantiles provide error bars for the predicted function values.}
\label{fig:gp_ex}
\end{figure}

Given the mean and covariance functions as well as the measurements, we can form the Gaussian process regressor. Assuming that the noises are independent of the function values, we can write the joint distribution of the observed values and the unknown function values as follows:
\begin{equation}
\begin{pmatrix}
  \mathbf{y} \\
  f(\mathbf{Z}^*) \\
\end{pmatrix}
\sim
\mathcal{N}\left( 
\begin{pmatrix}
  m(\mathbf{Z}) \\
  m(\mathbf{Z}^*) \\
\end{pmatrix},
\begin{pmatrix}
  \left( \mathbf{K} + \sigma_\mathrm{n}^2 I \right)
  & \mathbf{k}^\top(\mathbf{Z}^*) \\
  \mathbf{k}(\mathbf{Z}^*) & k(\mathbf{Z}^*,\mathbf{Z}^*)
\end{pmatrix}
\right),
\end{equation}
where $\mathbf{y} = \begin{pmatrix} y_1 & \ldots & y_N \end{pmatrix}^\top$, $m(\mathbf{Z}) = \begin{pmatrix} m(\mathbf{z}_1) & \cdots & m(\mathbf{z}_N) \end{pmatrix}^\top$, $m(\mathbf{Z}^*) = \begin{pmatrix} m(\mathbf{z}_1^*) & \ldots & m(\mathbf{z}_M^*) \end{pmatrix}^\top$, and $\mathbf{K}$ and $\mathbf{k}(\mathbf{Z}^*)$ denote matrices with element $(i,j)$ given as $k(\mathbf{z}_i,\mathbf{z}_j)$ and $k(\mathbf{z}_i^*,\mathbf{z}_j)$, respectively. By conditioning this joint Gaussian distribution on the measurements $\mathbf{y}$ we get that the conditional (i.e., posterior) distribution of the function values $f(\mathbf{Z}^*) = \begin{pmatrix} f(\mathbf{z}_1^*) & \ldots & f(\mathbf{z}_M^*) \end{pmatrix}^\top$ is Gaussian with the mean and covariance
\begin{equation}
\begin{split}
  \mathrm{E}[ f(\mathbf{Z}^*) \mid \mathbf{y} ]
  &= m(\mathbf{Z}^*) \\
  &+ \mathbf{k}(\mathbf{Z}^*) \,
    \left[ \mathbf{K} + \sigma_\mathrm{n}^2 \, \mathbf{I} \right]^{-1} \,
    (\mathbf{y} - m(\mathbf{Z})), \\
  \mathrm{Cov}[ f(\mathbf{Z}^*) \mid \mathbf{y} ]
  &= k(\mathbf{Z}^*,\mathbf{Z}^{*}) \\
  &- \mathbf{k}(\mathbf{Z}^*) \,
  \left[ \mathbf{K} + \sigma_\mathrm{n}^2 \, \mathbf{I} \right]^{-1} \,
  \mathbf{k}^\top(\mathbf{Z}^{*}).
\end{split}
  \label{eq:gp_reg}
\end{equation}
These are the fundamental equations of Gaussian process regression. An example of Gaussian process regression with squared exponential covariance function is shown in Figure~\ref{fig:gp_ex}.

\subsubsection{Hyperparameter learning}

Even though Gaussian process regression is a non-parametric method, for which we do not need to fix a parametric class of functions, the mean and covariance functions can have unknown hyperparameters $\boldsymbol{\varphi}$ which can be estimated from data. For example, the squared exponential covariance function in Equation~\eqref{eq:se_cov} has the hyperparameters $\boldsymbol{\varphi} = (s,\ell)$.

A common way to estimate the parameters is to maximize the marginal likelihood -- also called evidence -- $p(\mathbf{y} \mid \boldsymbol{\varphi})$  of the measurements, or equivalently, minimize the negative log-likelihood of the measurements
\begin{equation}
\begin{split}
  - \log p(\mathbf{y} \mid \boldsymbol{\varphi})
  &= \frac{1}{2}
  \log | 2 \pi \, (\mathbf{K}_{\boldsymbol{\varphi}} + \sigma_\mathrm{n}^2 \, \mathbf{I}) | \\
  &
  + \frac{1}{2} (\mathbf{y} - m_{\boldsymbol{\varphi}}(\mathbf{Z}))^\top \,
  \left[ \mathbf{K}_{\boldsymbol{\varphi}} + \sigma_\mathrm{n}^2 \, \mathbf{I} \right]^{-1} \\
  &\qquad \times (\mathbf{y} - m_{\boldsymbol{\varphi}}(\mathbf{Z})).
\end{split}
\label{eq:gp_lh}
\end{equation}
The gradient of this function with respect to the hyperparameters is also available \citep[see, e.g.,][]{Rasmussen+Williams:2006} which allows for the use of gradient-based optimization methods to estimate the parameters.

Instead of using the maximum likelihood method to estimate the parameters, it is also possible to use a Bayesian approach to the problem and consider the posterior distribution of the hyperparameters
\begin{equation}
  p( \boldsymbol{\varphi} \mid \mathbf{y}) = 
  \frac{p(\mathbf{y} \mid \boldsymbol{\varphi}) \, p(\boldsymbol{\varphi})}
  {\int p(\mathbf{y} \mid \boldsymbol{\varphi}) \,
    p(\boldsymbol{\varphi}) \, \mathrm{d}\boldsymbol{\varphi}},
\end{equation}
where $p(\boldsymbol{\varphi})$ is the prior distribution of the hyperparameters. We can, for example, compute the maximum a posteriori estimate of the parameters by finding the maximum of this distribution or use Markov chain Monte Carlo (MCMC) methods \citep{Brooks+Gelman+Jones+Meng:2011} to estimate the statistics of the distribution.

In what follows, to avoid notational clutter, we drop out the hyperparameters from the Gaussian process formulations and inference methods although they are commonly estimated as part of the Gaussian process learning.

\subsubsection{Reduction of computational complexity}

A limitation of Gaussian process regression in its explicit form is that the computational complexities of the regression Equations~\eqref{eq:gp_reg} and likelihood Equation~\eqref{eq:gp_lh} are cubic $\mathcal{O}(N^3)$ in the number of measurements $N$. This is due to the $N \times N$ matrix inversion appearing in the equations which, even when implemented with Cholesky or LU decompositions, needs a cubic number of computational steps.

Ways of solving the computational complexity problem are, for example, sparse approximations using inducing points \citep{Quinonero-Candela+Rasmussen:2005,Rasmussen+Williams:2006,Titsias:2009}, approximating the problem with a discrete Gaussian random field model \citep{Lindgren+Rue+Linstrom:2011}, or use of random or deterministic basis/spectral expansions \citep{Quinonero-Candela:2010,Solin+Sarkka:2018}. %

\subsection{GP-NFIR, GP-NARX, GP-NOE, and related models}

In system identification, we can use Gaussian processes to model unknown input-output relationships in time series. Several different model architectures are available for this purpose. Let us assume that we have a system with input sequence $u_1,u_2,\ldots$ and output sequence $y_1,y_2,\ldots$ and the aim is to predict the outputs from inputs. We also assume that we have been given a set of training data consisting of known inputs and (noisy) outputs. In the following, we present some typically used architectures that have been proposed for this purpose. More details can be found in the monograph of \citet{Kocijan:2016}.

\subsubsection{GP-NFIR model}

\begin{figure}[htb!]
\begin{center}
\begin{tikzpicture}[scale=1.0]
\draw[very thick] (1,0) -- (4,0) -- (4,2.5) -- (1,2.5) -- cycle;
\node[align=center] at (2.5,1.25) {\bf GP model};
\draw[thick,->] (0,0.5) node [anchor=east] {$u_{k-m}$} -- (1,0.5);
\draw[thick,->] (0,1.5) node [anchor=east] {$u_{k-1}$} -- (1,1.5);
\draw[thick,->] (0,2.0)  node [anchor=east] {$u_{k}$}  -- (1,2.0);
\draw[dashed] (0.5,1.25)  -- (0.5,0.75);
\draw[thick,->] (4,1.25) -- (5,1.25);
\draw[very thick] (5.25,1.25) circle [radius=0.25];
\draw[thick,->] (5.5,1.25) -- (6.5,1.25) node [anchor=west] {$y_{k}$};
\draw[thick,->] (5.25,2.25) node [anchor=south] {$\epsilon_{k}$} -- (5.25,1.5);
\end{tikzpicture}
\end{center}
\caption{In GP-NFIR model the Gaussian process regressor is used to predict next output from previous inputs.}
\label{fig:gp-nfir}
\end{figure}
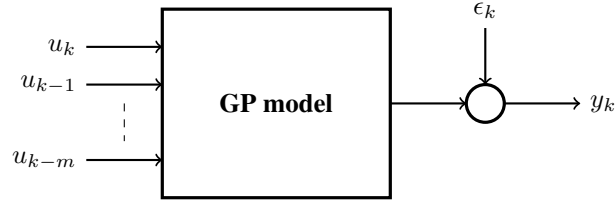

The Gaussian process non-linear finite impulse response (GP-NFIR) model \citep{Ackermann:2011,Kocijan:2016} has the form (see Figure~\ref{fig:gp-nfir})
\begin{equation}
  \hat{y}_k = f(u_{k-1},\ldots,u_{k-m}),
\end{equation}
where $f(\cdot)$ is an unknown mapping which we model as a Gaussian process, and $\hat{y}_k$ denotes the estimate produced by the regressor. In this model, we form a Gaussian process regressor that predicts the current output from a finite number of previous inputs. This model can be identified by reducing it into a Gaussian process regression model
\begin{subequations}
\begin{align}
  y_k &= f(\mathbf{z}_k) + \epsilon_k,
  \label{eq:std_gp_model}
\end{align}
\end{subequations}
where $\mathbf{z}_k = \begin{pmatrix} u_{k-1} & \ldots & u_{k-m} \end{pmatrix}^\top$ and by using standard Gaussian process regression methods on it. 

\subsubsection{GP-NARX model}

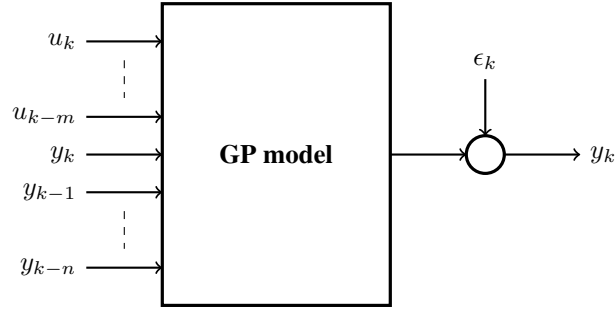
\begin{figure}[htb!]
\begin{center}
\begin{tikzpicture}[scale=1.0]
\draw[very thick] (1,0) -- (4,0) -- (4,4) -- (1,4) -- cycle;
\node[align=center] at (2.5,2) {\bf GP model};
\draw[thick,->] (0,0.5) node [anchor=east] {$y_{k-n}$} -- (1,0.5);
\draw[thick,->] (0,1.5) node [anchor=east] {$y_{k-1}$} -- (1,1.5);
\draw[thick,->] (0,2.0)  node [anchor=east] {$y_{k}$}  -- (1,2.0);
\draw[thick,->] (0,2.5) node [anchor=east] {$u_{k-m}$} -- (1,2.5);
\draw[dashed] (0.5,3.25)  -- (0.5,2.75);
\draw[thick,->] (0,3.5) node [anchor=east] {$u_{k}$} -- (1,3.5);
\draw[dashed] (0.5,1.25)  -- (0.5,0.75);
\draw[thick,->] (4,2) -- (5,2);
\draw[very thick] (5.25,2) circle [radius=0.25];
\draw[thick,->] (5.5,2) -- (6.5,2) node [anchor=west] {$y_{k}$};
\draw[thick,->] (5.25,3.0) node [anchor=south] {$\epsilon_{k}$} -- (5.25,2.25);
\end{tikzpicture}
\end{center}
\caption{In GP-NARX model the Gaussian process is used to predict the next output from the previous inputs and outputs.}
\label{fig:gp-narx}
\end{figure}

The Gaussian process nonlinear autoregressive model with exogenous input (GP-NARX) \citep{Kocijan:2005,Kocijan:2016} is a model of the form (see Figure~\ref{fig:gp-narx})
\begin{equation}
  y_k = f(y_{k-1},\ldots,y_{k-n},u_{k-1},\ldots,u_{k-m}) + \epsilon_k,
\end{equation}
where $\epsilon_k$ is a Gaussian random variable. This model can be reduced to a Gaussian process regression problem by setting $\mathbf{z}_k = \begin{pmatrix} y_{k-1} & \cdots & y_{k-n} & u_{k-1} & \cdots & u_{k-m} \end{pmatrix}^\top$ in Equation~\eqref{eq:std_gp_model}. 

\subsubsection{GP-NOE model}

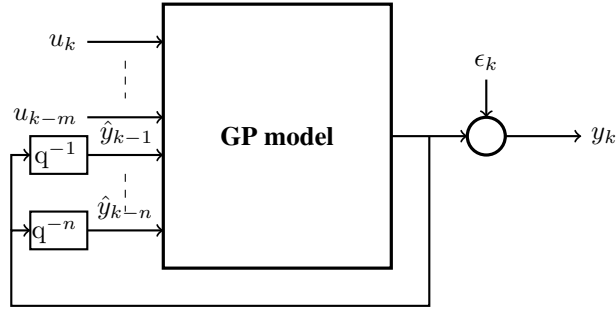
\begin{figure}[htb!]
\begin{center}
\begin{tikzpicture}[scale=1.0]
\begin{scope}[shift={(1,0)}]
\draw[very thick] (1,0.5) -- (4,0.5) -- (4,4) -- (1,4) -- cycle;
\node[align=center] at (2.5,2.25) {\bf GP model};
\draw[thick,->] (0,1.0) node [anchor=east] {$\mathrm{q}^{-n}$} -- (1,1.0)  node [anchor=south east] {$\hat{y}_{k-n}$};
\draw[thick] (-0.75,0.75) -- (0,0.75) -- (0,1.25) -- (-0.75,1.25) -- cycle;
\draw[thick,->] (0,2.0) node [anchor=east] {$\mathrm{q}^{-1}$} -- (1,2.0)  node [anchor=south east] {$\hat{y}_{k-1}$};
\draw[thick] (-0.75,1.75) -- (0,1.75) -- (0,2.25) -- (-0.75,2.25) -- cycle;
\draw[thick,->] (0,2.5) node [anchor=east] {$u_{k-m}$} -- (1,2.5);
\draw[dashed] (0.5,3.25)  -- (0.5,2.75);
\draw[thick,->] (0,3.5) node [anchor=east] {$u_{k}$} -- (1,3.5);
\draw[dashed] (0.5,1.75)  -- (0.5,1.25);
\draw[thick,->] (4,2.25) -- (5,2.25);
\draw[very thick] (5.25,2.25) circle [radius=0.25];
\draw[thick,->] (5.5,2.25) -- (6.5,2.25) node [anchor=west] {$y_{k}$};
\draw[thick,->] (5.25,3.0) node [anchor=south] {$\epsilon_{k}$} -- (5.25,2.5);
\draw[thick,->] (4.5,2.25) -- (4.5,0) -- (-1,0) -- (-1,1) -- (-0.75,1.0);
\draw[thick,->] (-1,1) -- (-1,2.0) -- (-0.75,2.0);
\end{scope}
\end{tikzpicture}
\end{center}
\caption{The GP-NOE model uses previous inputs and the previous outputs of the Gaussian process regressor to predict the next output. In the figure, $\mathrm{q}^{-n}$ denotes an $n$-step delay operator.}
\label{fig:gp-noe}
\end{figure}

In Gaussian process nonlinear output error (GP-NOE) model \citep{Kocijan:2011,Kocijan:2016} we form a Gaussian process regressor for the problem (see Figure~\ref{fig:gp-noe})
\begin{equation}
  y_k = f(\hat{y}_{k-1},\ldots,\hat{y}_{k-n},u_{k-1},\ldots,u_{k-m}) + \epsilon_k,
\end{equation}
where $\hat{y}_{k-1},\ldots,\hat{y}_{k-n}$ are the Gaussian process regressor predictions from the previous steps. 

Learning in this kind of model requires further approximations because the predictions of the Gaussian process are directly used as inputs on the next step.

\subsubsection{Other model architectures}

As discussed in \citet{Kocijan:2016}, it is also possible extend these architectures to, for example, GP-NARMAX (nonlinear autoregressive and moving average model with exogenous
input) models and NJB (nonlinear Box–Jenkins) models.

\subsection{Gaussian process state-space (GPSS) models}

Another approach to system identification is to form a state-space model where the dynamic and measurement models are identified using Gaussian process regression methods. This leads to so-called Gaussian process state-space models.

\subsubsection{General GPSS model}

\begin{figure}[htb!]
\begin{center}
\begin{tikzpicture}[scale=1.0]
\begin{scope}[shift={(1,3)}]
\draw[very thick] (1,0.5) -- (4,0.5) -- (4,2.5) -- (1,2.5) -- cycle;
\node[align=center] at (2.5,1.5) {\bf GP model $f$};
\draw[thick,->] (0,1.0) node [anchor=east] {$\mathrm{q}^{-1}$} -- (1,1.0) node [anchor=south east] {$\mathbf{x}_{k-1}$};
\draw[thick] (-0.75,0.75) -- (0,0.75) -- (0,1.25) -- (-0.75,1.25) -- cycle;
\draw[thick,->] (0,2.0)  node [anchor=east] {$u_{k-1}$}  -- (1,2.0);
\draw[thick,->] (4,1.5) -- (5,1.5);
\draw[very thick] (5.25,1.5) circle [radius=0.25];
\draw[thick,->] (5.5,1.5) -- (6.5,1.5) node [anchor=west] {$\mathbf{x}_{k}$};
\draw[thick,->] (5.25,2.5) node [anchor=south] {$\mathbf{w}_{k}$} -- (5.25,1.75);
\draw[thick,->] (6,1.5) -- (6,0) -- (-1,0) -- (-1,1) -- (-0.75,1.0);
\end{scope}
\draw[thick,->] (0,3) -- (0,2) -- (2,2) node [anchor=south east] {$\mathbf{x}_{k}$};
\begin{scope}[shift={(1,0)}]
\draw[very thick] (1,0.5) -- (4,0.5) -- (4,2.5) -- (1,2.5) -- cycle;
\node[align=center] at (2.5,1.5) {\bf GP model $g$};
\draw[thick,->] (0,1.0) node [anchor=east] {$u_k$} -- (1,1.0);
\draw[thick,->] (4,1.5) -- (5,1.5);
\draw[very thick] (5.25,1.5) circle [radius=0.25];
\draw[thick,->] (5.5,1.5) -- (6.5,1.5) node [anchor=west] {$y_{k}$};
\draw[thick,->] (5.25,2.5) node [anchor=south] {$\epsilon_{k}$} -- (5.25,1.75);
\end{scope}
\end{tikzpicture}
\end{center}
\caption{In GPSS model we learn a Gaussian process regressor for approximating the dynamic and measurement models in a state-space model. In this figure, $\mathrm{q}^{-1}$ denotes a one-step delay operator.}
\label{fig:gpss}
\end{figure}
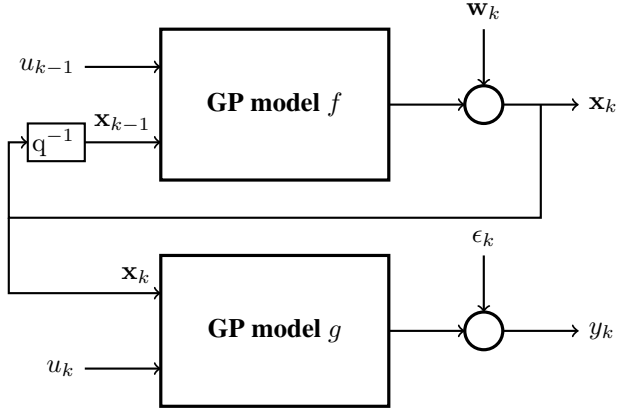

A Gaussian process state-space (GPSS) model (see Figure~\ref{fig:gpss}) has the mathematical form \citep[e.g.][]{Kocijan:2016}
\begin{subequations}
\begin{align}
  \mathbf{x}_{k+1} &= f(\mathbf{x}_k,u_k) + \mathbf{w}_k, \label{eq:ssmodel_fgua} \\
  y_k &= g(\mathbf{x}_k,u_k) + \epsilon_k, \label{eq:ssmodel_fgub}
\end{align}
\label{eq:ssmodel_fgu}
\end{subequations}
where the state vector $\mathbf{x}_k$, $k=0,1,2,\ldots,N$ contains the current state of the system,  $u_1,u_2,\ldots$ is the input sequence and $y_1,y_2,\ldots$ is the output sequence. In the model, $\mathbf{w}_k$ is a Gaussian distributed process noise. The aim is now to learn the functions $f(\mathbf{x}_k,u_k)$ and $g(\mathbf{x}_k,u_k)$, which are modeled as Gaussian processes, given the input and output sequences, or in some cases, also given direct observations of the state vector.

\subsubsection{Learning with fully observed state}

When the state vector $\mathbf{x}_k$ is fully observed, then both the dynamic model \eqref{eq:ssmodel_fgua} and measurement model \eqref{eq:ssmodel_fgub} become standard Gaussian process regression models. In dynamic model \eqref{eq:ssmodel_fgua}, the training set consists of measurements $\mathbf{x}_{k+1}$ with the corresponding inputs $(\mathbf{x}_k,u_k)$, and in measurement model \eqref{eq:ssmodel_fgub} the measurements are $y_k$ with the corresponding inputs $(\mathbf{x}_k,u_k)$. This kind of fully observed models are important in many applications such as robotics \citep{Deisenroth:2015}.

After conditioning on the training data, the functions $f$ and $g$ will still be Gaussian processes and their mean and covariance functions are given by (multivariate generalizations) of Equations~\eqref{eq:gp_reg}. State estimation in this kind of models has been considered by \citet{Ko:2009} and \citet{Deisenroth:2011}, and it turns out that it is possible to construct closed-form Gaussian approximation (moment matching) based filters and smoothers for these models \citep{Deisenroth:2011}. Control problems related to this kind of models have been considered, for example, by \citet{Deisenroth:2015}.

\subsubsection{Marginalization of the GP}
When the states $\mathbf{x}_k$ are not observed, then we need to treat both the states and the Gaussian processes as unknown. There are a few different ways to cope with the model in that case, and one approach is the marginalization approach of \citet{Frigola+Lindsten:2013,Frigola:2014,Frigola:2014b} and \citet{Frigola:2016}. First note that if we have a method to learn $f$, we can learn both $f$ and $g$ using a state-augmentation trick \citep{Frigola:2016}: we define an augmented state as $\tilde{\mathbf{x}} = \begin{pmatrix} \mathbf{x} &\gamma \end{pmatrix}^\top$, Gaussian process $h(\boldsymbol{\gamma},u) = \begin{pmatrix} f(\mathbf{x},u) & g(\mathbf{x},u) \end{pmatrix}^\top$, and augmented process noise $\tilde{\mathbf{w}} = \begin{pmatrix} \mathbf{w}_k &\mathbf{0} \end{pmatrix}^\top$, which reduces the model to
\begin{subequations}
\begin{align}
  \tilde{\mathbf{x}}_{k+1} &= h(\tilde{\mathbf{x}}_k,u_k) + \tilde{\mathbf{w}}_k, \\
  y_k &= \gamma_k + \epsilon_k.
\end{align}
\end{subequations}
This is now a model with an unknown dynamic model, but with a given linear Gaussian measurement model $p(y_k \mid \tilde{\mathbf{x}}_k,u_k) = \mathcal{N}(y_k \mid \gamma_k,\sigma^2_\mathrm{n})$.

Thus, without a loss of generality, we can focus on models with an unknown dynamic model $f$, and a known measurement model:
\begin{subequations}
\begin{align}
  \mathbf{x}_{k+1} &= f(\mathbf{x}_k,u_k) + \mathbf{w}_k, \label{eq:ssmodel_fpua} \\
  y_k &\sim p(y_k \mid \mathbf{x}_k,u_k). \label{eq:ssmodel_fpub}
\end{align}
\label{eq:ssmodel_fpu}
\end{subequations}
The aim is to learn the function $f(\mathbf{x})$ from a sequence of measurement data $y_k$. 

One way to understand \citep{Frigola:2016} this model is that we could hypothetically generate data from it by sampling the (infinite-dimensional) function $f(\mathbf{x},u)$, and then starting from $\mathbf{x}_0$ sequentially produce each $\{ \mathbf{f}_1,\mathbf{x}_1, \ldots,  \mathbf{f}_N,\mathbf{x}_N \}$, where we have denoted $\mathbf{f}_k = f(\mathbf{x}_{k-1},u_{k-1})$ for $k=1,2,\ldots,N$. Each of the conditional distributions
\begin{equation}
  p(\mathbf{f}_k \mid \mathbf{f}_{1:k-1}, \mathbf{x}_{0:k-1})
\end{equation}
turns out to be Gaussian. Note that above, we have introduced the short-hand notation $ \mathbf{f}_{1:k} = (\mathbf{f}_1,\ldots,\mathbf{f}_k)$ which we will use also in the rest of this article.

The above observation allows us to integrate out (i.e., marginalize) the Gaussian process from the model in closed form. The result is the following representation \citep{Frigola+Lindsten:2013}:
\begin{equation}
\begin{split}
  p(\mathbf{x}_{0:N})
  &= \prod_{k=1}^N \mathcal{N}(\mathbf{x}_k \mid \boldsymbol{\mu}_k(\mathbf{x}_{0:k-1}),
    \boldsymbol{\Sigma}_k(\mathbf{x}_{0:k-1})),
\end{split}
\label{eq:gpss_nonmarkov}
\end{equation}
where the means $\boldsymbol{\mu}_k(\mathbf{x}_{0:k-1})$ and covariances  $\boldsymbol{\Sigma}_k(\mathbf{x}_{0:k-1})$ are (quite complicated) functions of the prior mean and covariance functions of the Gaussian process $f$, which are evaluated on the whole previous state history. The above equation defines a non-Markovian prior model for the state sequence $\mathbf{x}_k$, $k=1,\ldots,N$. 

For a given $\mathbf{x}_{0:N}$, the distribution $p(f(\mathbf{x}^*) \mid \mathbf{x}^*, \mathbf{x}_{0:N})$ for a test point $\mathbf{x}^*$ can be computed by using conventional Gaussian process prediction equation as follow:
\begin{equation}
\begin{split}
  &p(f(\mathbf{x}^*) \mid \mathbf{x}^*, y_{1:N}) \\
  &= \int p(f(\mathbf{x}^*) \mid \mathbf{x}^*, \mathbf{x}_{0:N}) \,
      p(\mathbf{x}_{0:N} \mid y_{1:N}) \, \mathrm{d}\mathbf{x}_{0:N},
\end{split}
\end{equation}
which can be numerically approximated, provided that we use a convenient (e.g. Monte Carlo) approximation for $p(\mathbf{x}_{0:N} \mid y_{1:N})$.

Given the model \eqref{eq:gpss_nonmarkov} it is then possible to use, for example, particle Markov chain Monte Carlo methods \citep{Frigola+Lindsten:2013} to sample state trajectories from the posterior distribution $p(\mathbf{x}_{0:N} \mid y_{1:N})$ jointly with the parameters of the model, which provides a Monte Carlo approximation to the above integral. Other proposed approaches are, for example, particle stochastic approximation expectation--maximization \citep[EM,][]{Frigola:2014} which uses a Monte Carlo approximation to the EM algorithm aiming at computing the maximum likelihood estimates of the parameters while handling the states as missing data. %

\subsubsection{Approximation of the GP}
Another way to approach the problem where both the states and Gaussian processes are unknown is to approximate the Gaussian process as finite-dimensional parametric model and use conventional parameter estimation methods to the model. %

One possible approximation considered in \citet{Svensson:2016} is to employ a Karhunen--Loeve type of basis function expansion of the Gaussian process as follows:
\begin{equation}
  f(\mathbf{x},\mathbf{u}) = \sum_{i=1}^S c_i \, \phi_i(\mathbf{x},\mathbf{u}),
\end{equation}
where $\phi_i(\mathbf{x},\mathbf{u})$ are deterministic basis functions (e.g. sinusoidals) and $c_i$ are Gaussian random variables. With this approximation, the model in Equation~\eqref{eq:ssmodel_fpu} becomes
\begin{subequations}
\begin{align}
  \mathbf{x}_{k+1} &=  \sum_{i=1}^S c_i \, \phi_i(\mathbf{x}_k,\mathbf{u}_k) + \mathbf{w}_k, \label{eq:ssmodel_fpuas} \\
  y_k &\sim p(y_k \mid \mathbf{x}_k,u_k) \label{eq:ssmodel_fpubs},
\end{align}
\label{eq:ssmodel_fpus}
\end{subequations}
where learning of the Gaussian process $f$ reduces to estimation of the finite number of parameters $\mathbf{c} = \begin{pmatrix} c_1 & \cdots & c_S \end{pmatrix}^\top$ in the state-space model. The states and parameters in this model can now be determined using, for example, particle Markov chain Monte Carlo (PMCMC) methods \citep[see][]{Svensson:2016}.

Another possibility is to use inducing points (typically denoted with $\mathbf{u}$, but here we denote them with $\mathbf{f}^u$ to avoid confusion with the input sequence). In those approaches the idea is to first perform Gaussian process inference on the inducing points alone, that is, compute $p(\mathbf{f}^u \mid y_{1:N})$ and then compute the (approximate) predictions by conditioning on the inducing points instead of the original data. This leads to the approximation
\begin{equation}
\begin{split}
  &p(f(\mathbf{x}^*) \mid y_{1:N}) \\
  &= \int p(f(\mathbf{x}^*) \mid \mathbf{x}^*, \mathbf{x}_{0:N}) \,
      p(\mathbf{x}_{0:N} \mid y_{1:N}) \, \mathrm{d}\mathbf{x}_{0:N} \\
    &\approx \int p(f(\mathbf{x}^*) \mid \mathbf{x}^*, \mathbf{f}^u) \,
      p(\mathbf{f}^u \mid y_{1:N})
      \, \mathrm{d}\mathbf{f}^u.
\end{split}
\end{equation}
In the \citet{Turner:2010} the inducing points are learned using expectation--maximization (EM) algorithm. \citet{Frigola:2014b} propose a method for variational learning (or integration over) the inducing points by forming a variational Bayesian approximation $q(\mathbf{f}^u) \approx p(\mathbf{f}^u \mid y_{1:N})$, which further results in
\begin{equation}
\begin{split}
  &p(f(\mathbf{x}^*) \mid y_{1:N}) \\
    &\approx \int p(f(\mathbf{x}^*) \mid \mathbf{x}^*, \mathbf{f}^u) \,
      q(\mathbf{f}^u)
      \, \mathrm{d}\mathbf{f}^u
\end{split}
\end{equation}
and turns out to be analytically tractable as the optimal variational distribution $q(\mathbf{f}^u)$ is Gaussian.

\subsection{Spatio-temporal Gaussian process models}

\subsubsection{Temporal Gaussian processes}
Another way of modeling time series using Gaussian processes is by considering them as functions of time \citep{HartikainenSarkka2010} which are sampled at certain time instants $t_1,t_2,\ldots$:
\begin{equation}
\begin{split}
  f(t) &\sim \mathcal{GP}(m(t),k(t,t')), \\
  y_k &= f(t_k) + \epsilon_k.
\end{split}
\end{equation}
That is, instead of attempting to form a predictor from the previous measurements or inputs, the idea is to condition the temporal Gaussian process on its observed values and use the conditional Gaussian process for predicting values at new time points. 

Unfortunately, due to the cubic computational scaling of the Gaussian process regression, this quickly becomes in tractable when time series length increases. However, temporal Gaussian process regression is closely related to the classical Kalman filtering and Rauch-Tung-Striebel smoothing problems \citep[e.g.][]{Sarkka:2013}, which can be used to reduce the required computations. It turns out that provided that the Gaussian process is stationary, that is, the covariance function only depends on the time difference $k(t,t') = k(t-t')$, then, under certain restrictions, the Gaussian process regression problem is essentially equivalent to state-estimation in a model of the form
\begin{equation}
\begin{split}
  \frac{\mathrm{d}\mathbf{x}(t)}{\mathrm{d}t} &= \mathbf{A} \, \mathbf{x}(t)
  + \mathbf{B} \,\boldsymbol{\eta}(t), \\
  y_k &= \mathbf{C} \, \mathbf{x}(t_k) + \epsilon_k,
\end{split}
\end{equation}
where $\boldsymbol{\eta}(t)$ is a white noise process and the matrices $\mathbf{A}$, $\mathbf{B}$, and $\mathbf{C}$ are selected suitably to match the original covariance function. For example, the Mat\'ern covariance functions with half-integer smoothness parameters have exact representations as state-space models \citep{HartikainenSarkka2010}.

The Gaussian process regression problem can be now solved by applying a Kalman filter and Rauch-Tung-Striebel smoother on this problem. These methods have the fortunate property, that their complexity is linear $\mathcal{O}(N)$ with respect to the number of measurements $N$ as opposed to the cubic complexity of direct Gaussian process regression solution. %

\subsubsection{Spatio-temporal Gaussian processes}
A similar state-space approach also works for spatio-temporal Gaussian process models with a covariance function of a stationary form $k(\mathbf{z},t;\mathbf{z}',t') = k(\mathbf{z},\mathbf{z}';t-t')$. In that case, the state-estimation problem becomes infinite-dimensional, that is, a distributed parameter system
\begin{equation}
\begin{split}
  \frac{\mathrm{d}\mathbf{x}(\mathbf{z},t)}{\mathrm{d}t} &= \mathcal{A} \, \mathbf{x}(\mathbf{z},t)
  + \mathbf{B} \,\boldsymbol{\eta}(\mathbf{z},t), \\
  y_k &= \mathcal{C} \, \mathbf{x}(\mathbf{z},t_k) + \epsilon_k,
\end{split}
\end{equation}
where $\mathcal{A}$ is a matrix of operators and $\mathcal{C}$ is a matrix of functionals \citep{Sarkka+Solin+Hartikainen:2013}. The solution of the Gaussian process regression problem in this form requires the use of methods from partial differential equations, but in many cases we can obtain an exact $\mathcal{O}(N)$ inference procedure from this route.

\subsubsection{Latent force models}

In so-called latent force models \citep[][]{Alvarez+Luengo+Lawrence:2013} the idea is to infer latent force $\xi(t)$ in a differential equation model such as
\begin{equation}
\begin{split}
  \frac{\mathrm{d}^2 x(t)}{\mathrm{d}t^2}
  + \gamma \, \frac{\mathrm{d} x(t)}{\mathrm{d}t}
  + \nu^2  = \xi(t), 
\end{split}
\end{equation}
where $\xi(t)$ is an unknown function which is modeled as a Gaussian process. The inference in this model can be recast as Gaussian process regression with a modified covariance function. The idea can be further generalized to partial differential equation models and non-linear models when approximation methods such as Laplace approximation are used.

The inference in this kind of models can be further re-stated in the state-space form \citep{Sarkka:2019}, which also allows for the study of control problems on latent force models. This formulation also allows the analysis of observability and controllability properties of latent force models.

\section{Summary and Future Directions}

In this article, we have briefly outlined the main directions in system identification using Gaussian processes. The methods can be divided into three classes: (1) GP-NFIR, GP-NARX, and GP-NOE type of models which directly aim at learning the function from the previous inputs and outputs to the current output, (2) Gaussian process state-space models which aim to learn the dynamic and measurement models in the state-space model, and (3) Gaussian process regression methods for the spatio-temporal time series by direct or state-space Gaussian process regression. 

Active problems in research in this area appear to be, for example, joint learning and control in all the model types outlined here. Another direction of further study is the analysis of observability, identifiability, and controllability. In practice, the Gaussian process-based system identification methods are very similar to classical methods, and hence they can be expected to inherit many limitations and theoretical properties of the classical methods. 

An emerging research area in Gaussian process-based models are so-called deep Gaussian processes \citep{Damianou2013} which borrow the idea of deep neural networks by forming hierarchies of Gaussian processes. This kind of models could also turn out to be useful in system identification. Furthermore, Gaussian processes are also easily combined with first-principles-models (cf. latent force models described above) which allows for flexible gray box modeling using Gaussian processes.

One of the main obstacles in Gaussian process regression is still the computational scaling in the number of measurements, which is also inherited by all the new developments. Although several good approaches to tackle this problem have been proposed, the problem still is that they inherently replace the original model with an approximation. New better approaches to this problem are likely to appear in the near future. 

\section{Cross References}

\bibliographystyle{spbasic}
\bibliography{bib-gp-sysid}

\end{document}